# An end-to-end CNN framework for polarimetric vision tasks based on polarization-parameter-constructing network

**YONG WANG,[1,2,*] QI LIU,[3] HONGYU ZU,[1,2] XIAO LIU,[4] RUICHAO XIE,[1] AND FENG WANG[1,2]**

[1]Department of Information Engineering, Army Academy of Artillery and Air Defense, PLA, Hefei, 230031, China
[2]Anhui Provincial Key Laboratory of Polarization Imaging Detection Technology, Hefei,230031, China
[3]School of Internet, Anhui University, Hefei,230031, China
[4]Anhui Institute of Optics and Fine Mechanics, Chinese Academy of Sciences, Hefei,230031, China
*wy1983@outlook.com

**Abstract:** Pixel-wise operations between polarimetric images are important for processing polarization information. For the lack of such operations, the polarization information cannot be fully utilized in convolutional neural network(CNN). In this paper, a novel end-to-end CNN framework for polarization vision tasks is proposed, which enables the networks to take full advantage of polarimetric images. The framework consists of two sub-networks: a polarization-parameter-constructing network (PPCN) and a task network. PPCN implements pixel-wise operations between images in the CNN form with $1\times1$ convolution kernels. It takes raw polarimetric images as input, and outputs polarization-parametric images to task network so as to complete a vison task. By training together, the PPCN can learn to provide the most suitable polarization-parametric images for the task network and the dataset. Taking faster R-CNN as task network, the experimental results show that compared with existing methods, the proposed framework achieves much higher mean-average-precision (mAP) in object detection task.

## 1. Introduction

Polarimetric images can provide extra information about objects. Nevertheless, it is difficult to be fully utilized in computer vision tasks, since the polarimetric characteristics are changing so constantly that it is hard to model in open environments. To solve this problem, researchers introduced CNN-based methods[1-5], which have been proved very effective in modeling and tracing invariants and high-order features from complex changes. In previous studies, the inputs of CNNs are usually man-selected polarization-parametric images (raw polarimetric images captured by sensors, Stokes parametric images, DoLP, AoP, etc.), or the fusion results of them. However, few studies have focused on two questions: 1) Whether these particular inputs are the best choice for the objects; 2) Whether the networks can take full advantage of the polarimetric information.

Our opinion is:

1) Those inputs may not be the best choice. First, a small number of man-selected polarization parameters may not be able to describe the polarimetric characteristics of multi-objects fully and precisely. Recently, a detection network using polarimetric images as input was studied[3]. The experimental result shows that detection performances on different types of target are quite different though the network uses the same type of input polarimetric images. Second, most of the input parameters are not invariable. DoLP, for example, a very frequently-used parameter, is sensitive to imaging conditions.

2) Current vision task networks, such as Faster R-CNN[6] and ResNet[7], cannot make full use of polarimetric information. Pixel-wise operations between raw polarimetric images are an essential way to extract polarimetric information and construct polarization parameters. To implement such operations with CNN, $1\times1$ pixel-size convolution kernel is required. However, existing CNNs for vision tasks rarely use the $1\times1$ kernel except few situations[8]. This is because the networks are designed to extract spatial visual features, and $3\times3$ pixel-size or larger kernels are required in this case. Theoretically, through training, big kernels can be converted into kernels with all zero weights except the central point, which are equivalent to $1\times1$ ones. But back propagation will avoid such conversions, because this kind of conversion will degrade the feature-extracting capacity and the task performance.

This paper proposed a two-step, end-to-end CNN framework for polarimetric vision tasks, which was composed of two sub-networks: polarimetric-parameter-constructing network (PPCN) and task network. PPCN is a CNN designed to implement pixel-wise operations between raw polarimetric images and output a set of new images for the task network. The task network can be any type of vision task CNNs such as object detection network and classification networks. PPCN gives the network the ability to create different polarization-expressions for different objects and to extract polarimetric characteristics. The two sub-networks are connected and trained together. Through joint training, the PPCN can learn a model and produce a set of polarization-parametric images to maximize the performance of the task network. The structure of PPCN is scalable, which depends on computing resources or efficiency requirements. We also discussed and analyzed the hyper-parameters of PPCN. In addition, we built a dataset containing 3000 sets of raw polarimetric images and RGB images captured in open environments. Taking Faster R-CNN as a task network, we verified the framework on the dataset and achieved the highest performance among several polarimetric images input strategies.

## 2. Related Works

Typically, computer vision tasks, e.g., object detection and classification, can benefit from polarimetric information in three ways:

***Methods based on physical models.*** Such methods usually use a specific physical model or polarimetric characteristic model to distinguish targets from background[9-12]. Methods of this kind have good interpretability and physical significance, and usually achieve good performance. However, such methods are often with strict physical constraints, which generally cannot be satisfied in many open environments.

***Methods based on data fusion.*** In these methods, polarimetric images are fused with other images(polarimetric or non-polarimetric), to generate new images, which are expected to enrich the target detail or increase the contrast between target and background [13-17]. Information fusion theories are considered rather than physics in these methods. Flexibility and diversity are the advantages of data fusion theory. Nevertheless, such methods are not robust in many open environments, because the changes of characteristic, which is caused by instability of imaging conditions, is usually not considered. Moreover, different objects often lead to different fusion algorithms, which make multi-objects vision tasks to be hard to implement.

***Methods based on CNN.*** The CNN methods were introduced to solve the robustness problem in open environments. Taking Stokes parameters ($S_0$, $S_1$, and $S_2$) as input, a convolutional neural network with both 3-D and 2-D convolution layers is studied to distinguish different natural backgrounds[1]. Taking raw polarimetric images at $0^\circ$ ,$45^\circ$ and $90^\circ$ as input for RetinaNet[18], object detection performance is tested [3]. The detection average precision (AP) for cars is 0.90, oppositely, 0.38 for pedestrians. Another work attempts to improve detection performance in severe weather by using CNN and polarimetric images[5]. The object detection approach of infrared polarimetric images are also studied by using CNN[2, 4]. The demosaicing network for microgrid polarimeter imagery [19, 20]is another related work.

Methods based on physical model or data fusion often perform pixel-wise operations between polarimetric images. These operations are the process of polarization information fusing and reconstructing in physics, and can dig out more information from the input polarimetric images. Nevertheless, such methods are not robust in open environments. CNN-based methods are much more robust, but it is unable for these methods to take full advantage of polarimetric images due to the lack of pixel-wise operations.

We implemented pixel-wise operations in the form of CNN. By connecting to vision task networks, this module enabled existing networks to take full advantage of polarimetric images and learn the best polarimetric expressions for targets. Experimental results proved that our methods can achieve better performance.

## 3. The proposed method

### *3.1 Polarization-parameter-constructing network (PPCN)*

In order to obtain linear polarization information, a set of raw polarimetric images(e.g. , $I_0$, $I_{45}$, $I_{90}$, $I_{135}$)with multiple polarization directions (0°, 45°, 90°, 135°) need to be captured. Normally, these images are spatially aligned. $\left(I_0^{x,y}, I_{45}^{x,y}, I_{90}^{x,y}, I_{135}^{x,y}\right)$ is the values at coordinates ($x$, $y$) on this set of images. Perform an operation $f(\cdot)$ on $\left(I_0^{x,y}, I_{45}^{x,y}, I_{90}^{x,y}, I_{135}^{x,y}\right)$, get $P_f^{x,y}$.

$$P_f^{x,y} = f\left(I_0^{x,y}, I_{45}^{x,y}, I_{90}^{x,y}, I_{135}^{x,y}\right) \tag{1}$$

$P_f^{x,y}$ is a linear polarization parameter at ($x$, $y$). Perform $f(\cdot)$ at all coordinates of the image, we can get a linear polarization-parametric image $P_f$. Theoretically, linear polarization parameters can be constructed by defining different $f(\cdot)$, and can describe polarization characteristics of objects in different ways. Commonly-used parameters are Stokes parameters ($S_0$, $S_1$, $S_2$, $S_3$), angle of polarization (*AoP*), and degree of linear polarization (*DoLP*). Some creatures can even construct polarization-parametric images which are most suitable for their living environments by their biological vision system[21].

The polarization-parameter constructing process defined in Equation 1 can be implemented with a multi-layer artificial neural network(Fig.1). This network defines a non-linear mapping from input to output. It contains: 1) an input layer, which has four nodes corresponding to $\left(I_0^{x,y}, I_{45}^{x,y}, I_{90}^{x,y}, I_{135}^{x,y}\right)$ in this example; 2) a certain number of hidden layers; 3) an output layer, each node of whose corresponds to a certain polarization parameter. Theoretically, if the ANN is deep enough, it can fit any complex mapping model.

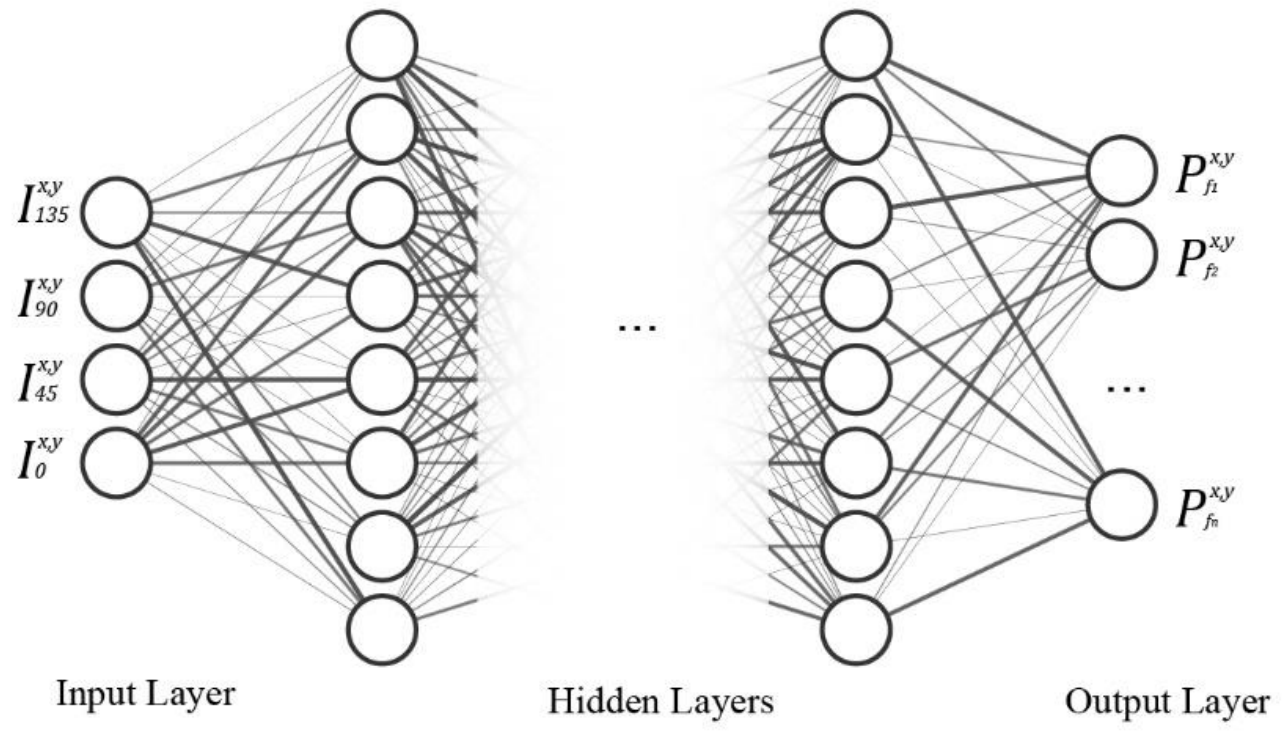

Fig. 1. An ANN for polarization-parameter constructing.

For images, the above process can be implemented with a 2-D convolutional neural network (Fig. 2).

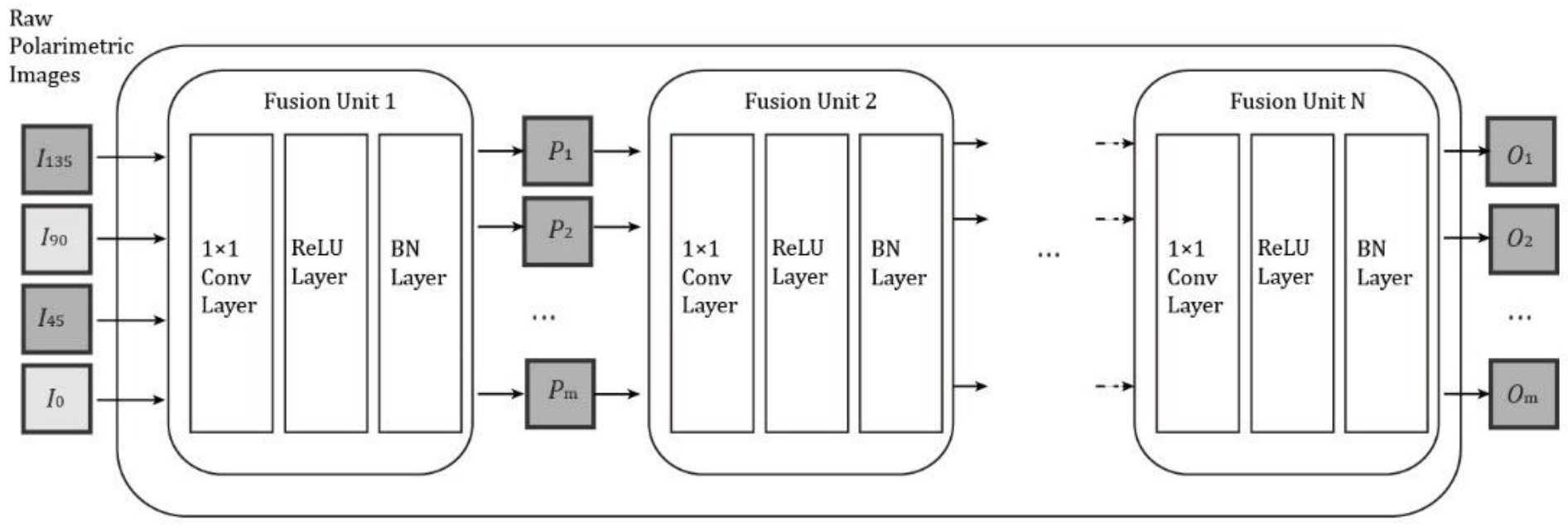

Fig. 2. Structure of PPCN.

This convolutional neural network for constructing polarization-parametric images is called polarization-parameter-constructing network (PPCN). It consists of: 1) an input unit, which provides a set of raw polarimetric images to the network; 2) multiple fusion units, each of which performs non-linear, pixel-wise operation between input images, and outputs a set of new images. The output of the previous unit is the input of the next; 3) an output unit, which provides the final polarization-parametric images produced by fusion units. Due to the 1×1 convolution kernel, the size of the output polarization-parameter images is consistent with the input images.

Each fusion unit contains a 1 × 1 convolution layer, a rectified linear unit(ReLU) layer[22] and a batch normalization(BN) layer [23]. 1×1 convolution layer implements pixel-wise linear operations between images, ReLU layer implements non-linear operations, and BN layer makes the network easier to converge.

For ease of expression, we use the output-image number of every unit to represent the structure of PPCN. For example, 4-8-16-8-5 indicates that the network consists of: 1) 1 input unit taking 4 raw polarimetric images as input; 2) 3 fusion units, output-image numbers of which are 8, 16 and 8; 3) 1 output unit which finally produces 5 different polarization-parametric images.

The structure of PPCN is a hyperparameter of the network, which needs to be manually selected.

### *3.2 PPCN structure hyperparameter*

To study the effect of structure hyperparameter on network performance, we constructed PPCNs using different hyperparameters and trained them to fit $S_0$, DoLP and AoP, whose definitions are shown below:

$$S_0 = I_0 + I_{90} \tag{2}$$

$$DoLP = \frac{\sqrt{\left(I_0 - I_{90}\right)^2 + \left(I_{45} - I_{135}\right)^2}}{S_0} \tag{3}$$

$$AoP = \frac{1}{2}\arctan\left(\frac{I_0 - I_{90}}{I_{45} - I_{135}}\right) \tag{4}$$

$S_0$ indicates a simple linear operation between images, DoLP represents a non-linear operation, and AoP contains an inverse trigonometric function. The computational complexity of the three parameters are significantly different, which is good for performance testing.

Experimental dataset contains 200 sets of polarized image samples. Normalized results of $S_0$, DoLP and AoP parameter images of each set are taken as ground truth.

Define loss function as below:

$$Loss = \frac{1}{N}\sum_{n=1}^{N}\left[\frac{1}{W \cdot H}\left(\left\|\hat{Y}_{S_0}^{(n)} - Y_{S_0}^{(n)}\right\|_2 + \left\|\hat{Y}_{DoLP}^{(n)} - Y_{DoLP}^{(n)}\right\|_2 + \left\|\hat{Y}_{AoP}^{(n)} - Y_{AoP}^{(n)}\right\|_2\right)\right] \tag{5}$$

Where $\|\cdot\|_2$ represents $l_2$ norm, $N$ is the total number of samples; $n$ is the sample index; $W$ and $H$ are the width and the height of the image; $\hat{Y}_{S_0}$, $\hat{Y}_{DoLP}$ and $\hat{Y}_{AoP}$ are the polarization parameter images output by the network, and $\hat{Y}_{S_0}$, $\hat{Y}_{DoLP}$ and $\hat{Y}_{AoP}$ are normalized results of $S_0$, DoLP and AoP according to their respective ranges. The normalization is used for avoiding the imbalance of the loss function caused by different ranges.

With training, the loss will decrease continually, as $\hat{Y}_{S_0}$, $\hat{Y}_{DoLP}$ and $\hat{Y}_{AoP}$ produced by PPCN are approximating to $Y_{S_0}$, $Y_{DoLP}$ and $Y_{AoP}$. Correspondingly, the stronger the fitting ability of the network, the smaller the loss value. After 200 epochs, the results are shown in Table 1.

**Table 1. Training Result with different structures**[a]

| Structures | Parameter count | GPU Memory Usage (MB) | Time Spent | Loss |
|---|---|---|---|---|
| 4-8-16-8-3 | 347 | 1453 | 4h 20m 52s | $1.16\times10^{-2}$ |
| 4-16-8-8-3 | 347 | 1453 | 4h 33m 18s | $9.42\times10^{-3}$ |
| 4-48-96-32-3 | 8,147 | 3817 | 4h 28m 19s | $7.29\times10^{-3}$ |
| 4-96-48-32-3 | 8,147 | 3817 | 4h 47m 34s | $5.85\times10^{-3}$ |
| 4-96-128-64-32-3 | 23,331 | 6109 | 5h 18m 31s | $5.87\times10^{-3}$ |
| 4-128-96-48-32-3 | 19,347 | 5705 | 5h 37m 48s | $4.21\times10^{-3}$ |

[a]GPU=1×1080ti，batch size= 2， learning rate=0.003， input resolution=1280×1024 pixel

Figure 3 shows a comparison of the fitting results with two different networks.

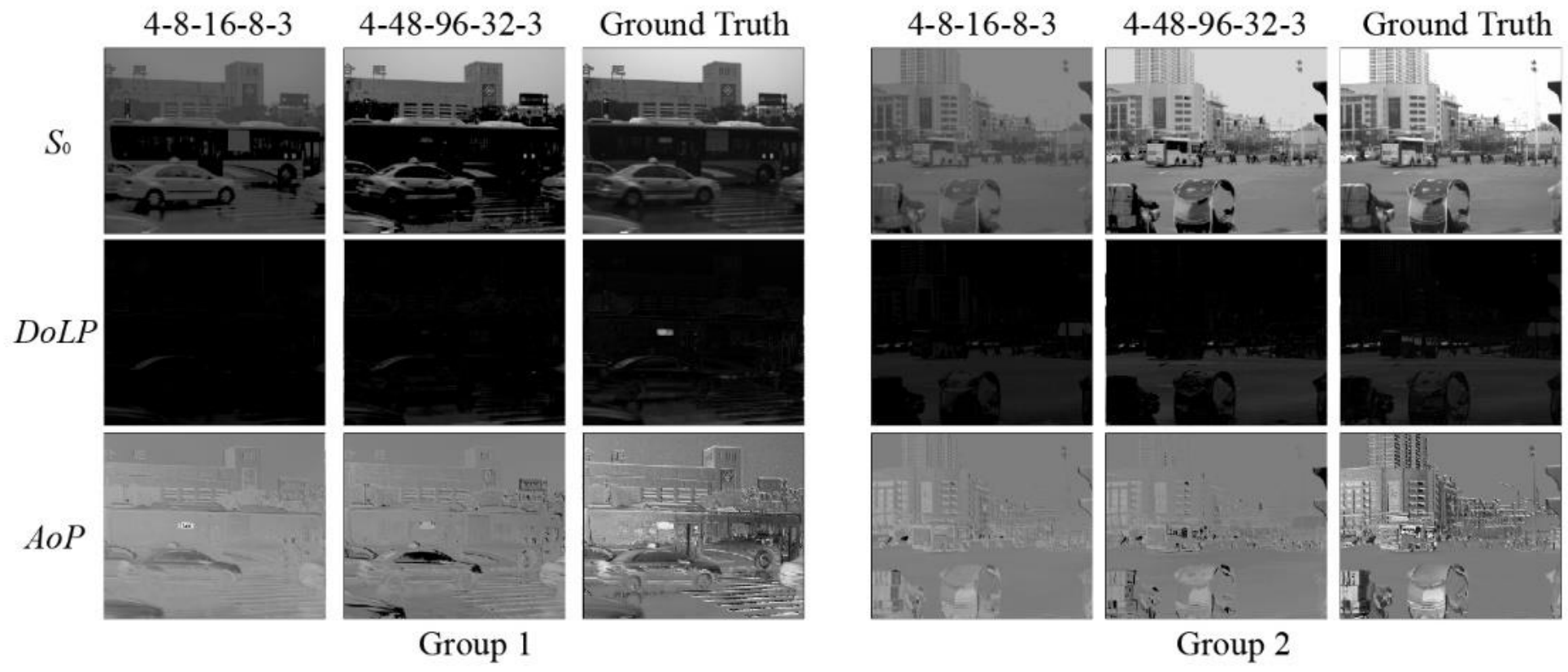

Fig. 3. Fusing Result on Testing Images.

Experimental result proves that larger network will lead to higher fitting accuracy, and more resource consumption. Time spent on training is very close, but much more GPU RAMs are needed when using lager networks.

### 3.3 Framework for vision task

PPCN cannot complete vison task, so a task network is needed. The task network can be classification networks(e.g. , VGGNet[24] and ResNet[7]), detection networks(e.g. , Faster R-CNN[6] and Yolo[25]) ,image fusion networks(e.g. , FusionCNN[26]), etc. It takes the output of PPCN as input to complete vison task. More importantly, by training together, the PPCN model will be optimized by the task network, so as to produce the most suitable polarization-parametric images and achieve the best performance.

PPCN and task network form an end-to-end CNN framework for polarimetric vison tasks (Fig. 4). It can take raw polarimetric images directly from imaging system, and then complete the task without any manual intervention.

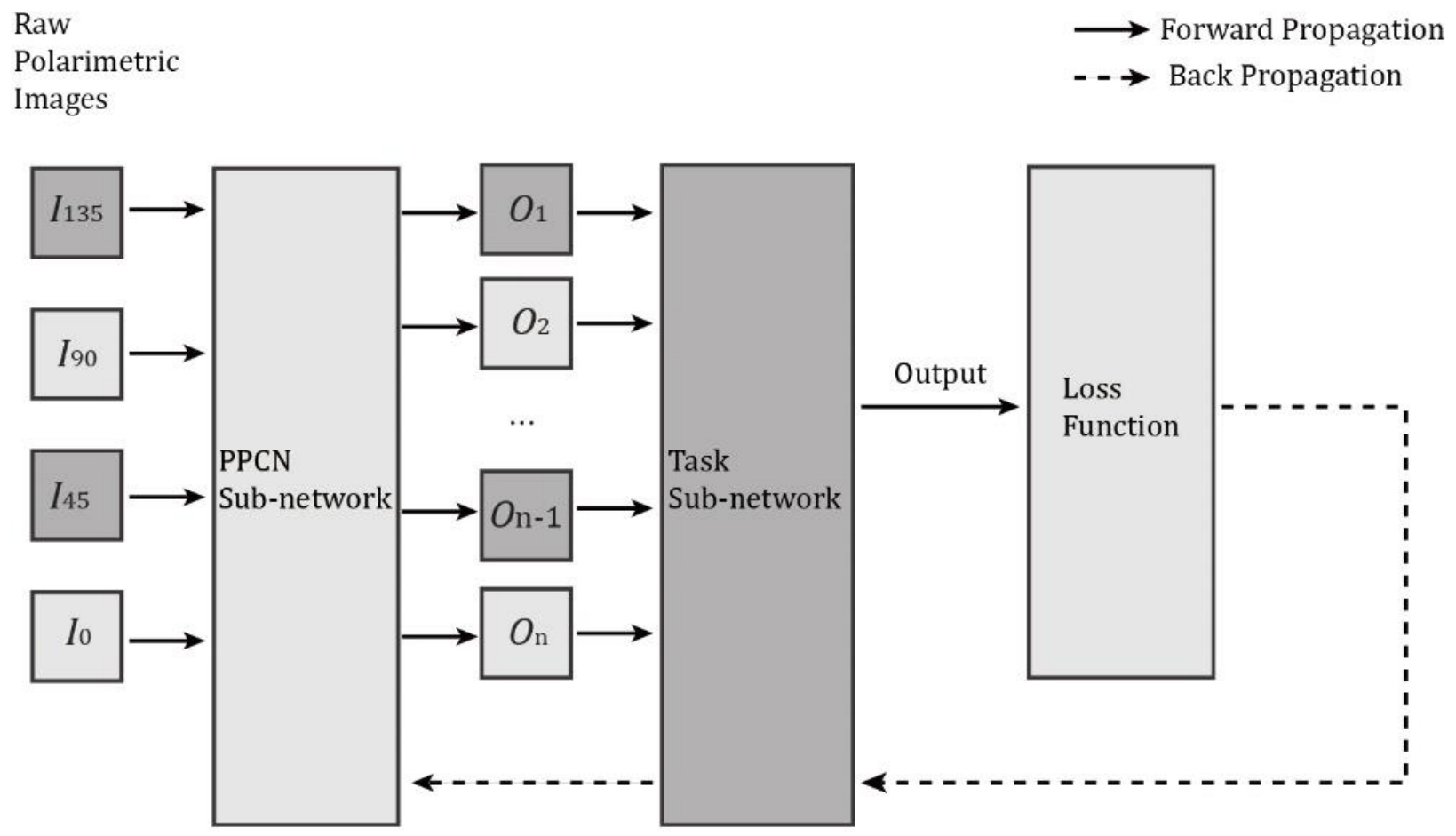

Fig. 4. Structure of framework with PPCN and task network.

During forward propagation, PPCN takes raw polarimetric images as input, performs a pixel-wise mapping, outputs polarization-parametric images, and then passes the images to the task network to complete vison task.

During back propagation, the output of the task network will be passed to loss function to get loss and calculate the gradient. Gradient will be transmitted back to the task network and PPCN according to the chain rule, and optimize the parameters of the two sub-networks.

These two processes are iterated until the performance of the task network is optimal or reaches a preset stop condition. Then the polarization parameter image output by PPCN can be considered to be the most suitable for the task network and the dataset.

## 4. Experiments

Taking Faster R-CNN as task network, we verified the proposed framework. A dataset consisting of polarization targets was built and a series of experiments were performed on it. First, as a hyperparameter, the number of polarization-parametric images produced by PPCN was selected carefully. Also, we visualized the polarization-parametric images produced by PPCN and tried to figure out how the PPCN extract the polarimetric characteristics of targets. At the same time, the task performance was compared between our framework and some other input strategies used in previous studies. We also made a simple comparison with the performance of network taking RGB images as input.

### 4.1 Dataset

Because there are few polarization image datasets taken in open environments for general vision tasks at present, a data set which contains 3000 sets of images was built. Each set of images consists of a RGB image and four raw polarimetric images with orientations of 0°, 45°, 90° and 135°, all of which are 1280 ×1024 pixels. Cars and pedestrians were labeled as targets with bounding boxes. Polarization characteristics of these two different categories of object are so different that we can evaluate the performance of the model more comprehensively. The composition and division of the data set are shown in Table 2.

**Table 2. composition and division of the data set**

| | Images Sets | Labeled | Pedestrians | Cars |
|---|---|---|---|---|
| Train Set | 2700 | 15051 | 6913 | 8138 |
| Val Set | 300 | 1427 | 712 | 715 |
| All | 3000 | 16478 | 7625 | 8853 |

To ensure the diversity of samples, images are captured from different scenes, orientations, distance and time periods. Target size and posture are fully taken into account, too. Some of the samples are shown in Fig. 5.

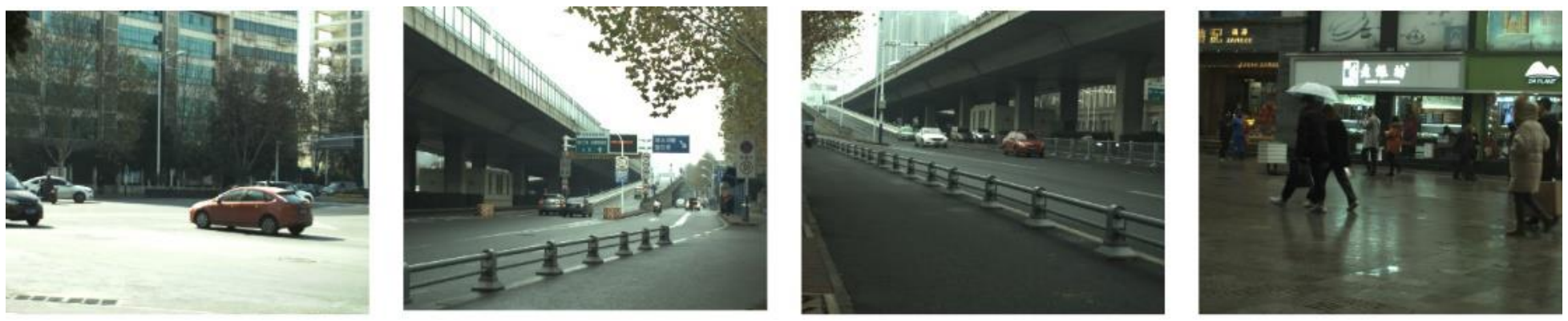

(a) (b) (c) (d)

Fig. 5. Different scenes in dataset. (a) Crossroads at 14:00am, with multi view of targets at different distance. (b)Entrance of overpass at 10:00am, with rear view of targets at different distance. (c)Exit of overpass at 17:00pm, with front view of targets at different distance. (d) Pedestrians on street.

### 4.2 Number of PPCN output

According to the limit of GPU RAM size, the PPCN structure was set to 4-48-96-32-16-x, where x represents the number of polarization-parametric images output by PPCN. We tested performance when x was 5, 7, 9 and 11. To evaluate the overall performance of the network for all types of targets, mAP was adopted as a performance criterion, while average precision (AP) was used to evaluate the performance for single type of target. Table 3 shows the results.

**Table 3. Detection Results Using Different Numbers of PPCN Output Number (for all targets)** [a]

| FPF-Net Structure | Backbone | mAP (%) |
|---|---|---|
| 4-48-96-32-16-1 | ResNet50 | 0.661 |
| 4-48-96-32-16-3 | ResNet50 | 0778 |
| 4-48-96-32-16-5 | ResNet50 | 0.782 |
| 4-48-96-32-16-7 | ResNet50 | 0.800 |
| 4-48-96-32-16-9 | ResNet50 | **0.827** |
| 4-48-96-32-16-11 | ResNet50 | 0.812 |

[a] GPU=4×1080ti, batch size= 2, learning rate = 0.01, input resolution=1280×1024 pixel

Experimental results show that the output of more than 9 images can no longer improve the performance on this dataset, indicating that 9 different polarimetric parameters learned from the dataset can provide sufficient polarization information about the targets and backgrounds.

Since the polarization characteristics of cars and pedestrians are significantly different, we could assume that the PPCN may need more output parameters to distinguish them. To prove that, we made a sub-dataset including only cars from the original dataset, and tested on it. The results are as follows.

**Table 4. Detection Results Using Different Numbers of PPCN Output Number(only for car)** [a]

| FPF-Net Structure | Backbone | AP (%) |
|---|---|---|
| 4-48-96-32-16-1 | ResNet50 | 0.830 |
| 4-48-96-32-16-3 | ResNet50 | 0.896 |
| 4-48-96-32-16-5 | ResNet50 | **0.915** |
| 4-48-96-32-16-7 | ResNet50 | **0.915** |
| 4-48-96-32-16-9 | ResNet50 | 0.913 |

[a]GPU=4×1080ti，batch size= 2 per GPU， learning rate = 0.01， input resolution=1280×1024 pixel.

As be seen from the table above, if only cars are concerned, 5 different parameters are enough to describe the polarimetric characteristics. The results prove that in order to fully describe two types of targets and backgrounds with significantly different polarization characteristics, PPCN has learned different polarization parameters for each of them.

### *4.3 Visualization of output images*

To figure out what PPCN learn from the data, we visualized the output images. The structure of the PPCN was set to 4-48-96-32-16-5. To avoid confusion caused by multi-objects, we trained the network on a sub-dataset including only cars. The outputs are mapped to grey images (Fig. 6).

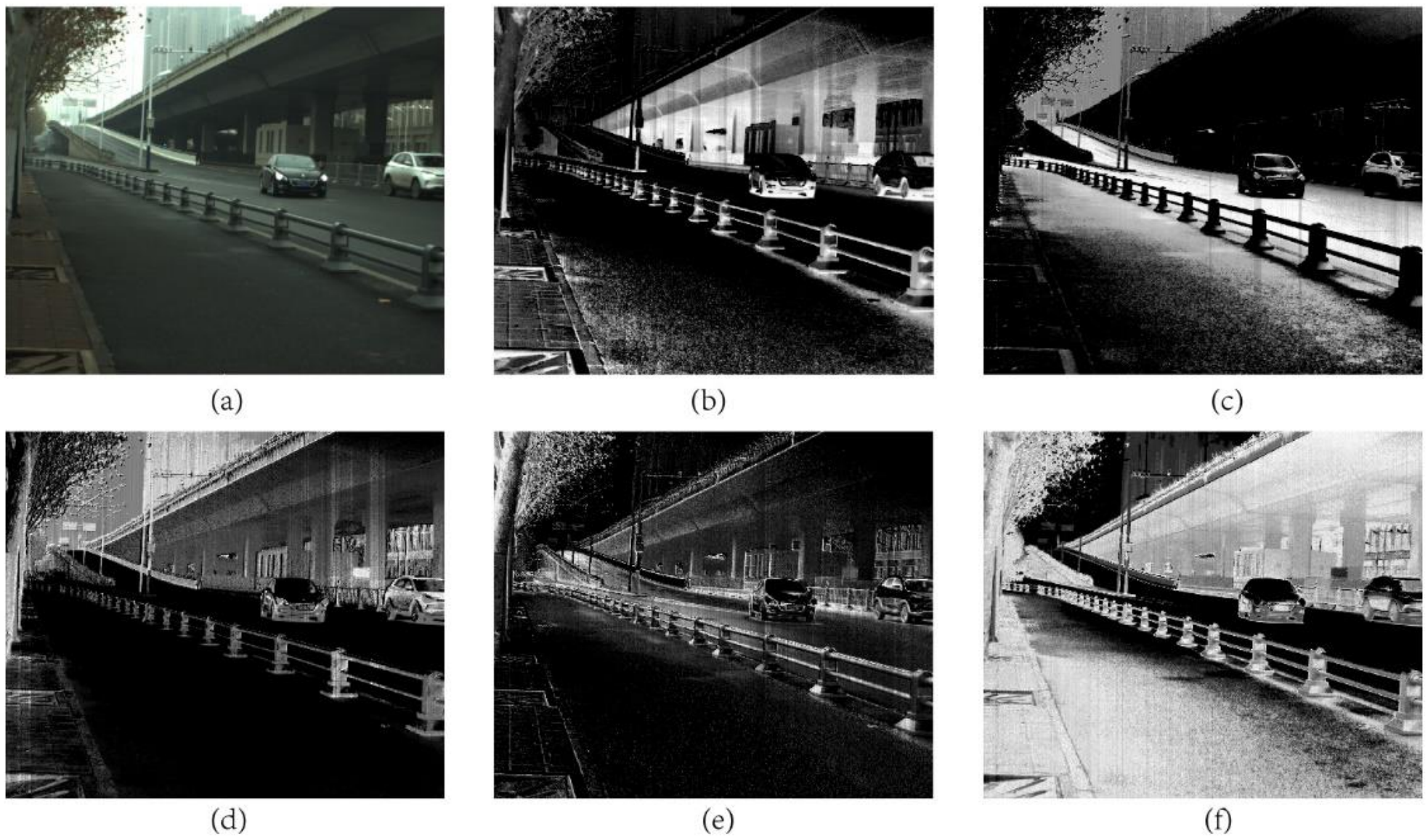

Fig. 6. Visualized PPCN output images of a sample. (a)RGB image of the sample. (b)-(f) PPCN outputs mapped to grey images.

The sub-dataset mainly consists of road scenes, which normally includes vehicles, roads, buildings and plants. According to Fig. 6, we could learn how the PPCN responds to different components of the scene.

1) Output images are quite different from each other, and information redundancy is not serious, which means that PPCN can describe objects in many ways. Taking the cars for example, different parts of the two cars are activated in different output images.
2) Details are enriched. In the case of the tree on the left, much more details are revealed in (d)-(f), and we can tell different parts of it.
3) Backgrounds are strongly suppressed in different images, but the targets(cars) are always activated. Roads are suppressed in (d). Trees and the overpass are suppressed in (c). Buildings are suppressed in (e). But in all the images, cars are activated in whole or in part.

We can assume that PPCN tries to model targets and backgrounds in more than one way. Considering that PPCN achieves this by pixel-wise operations only, it can be assumed that polarization information is beneficial to feature expression.

### 4.4 Performance comparison

We tested the performance using PPCN, RGB images, raw polarimetric images, AoP images, DoLP images, $S_0$ images and their combinations as input respectively in Fig. 7.

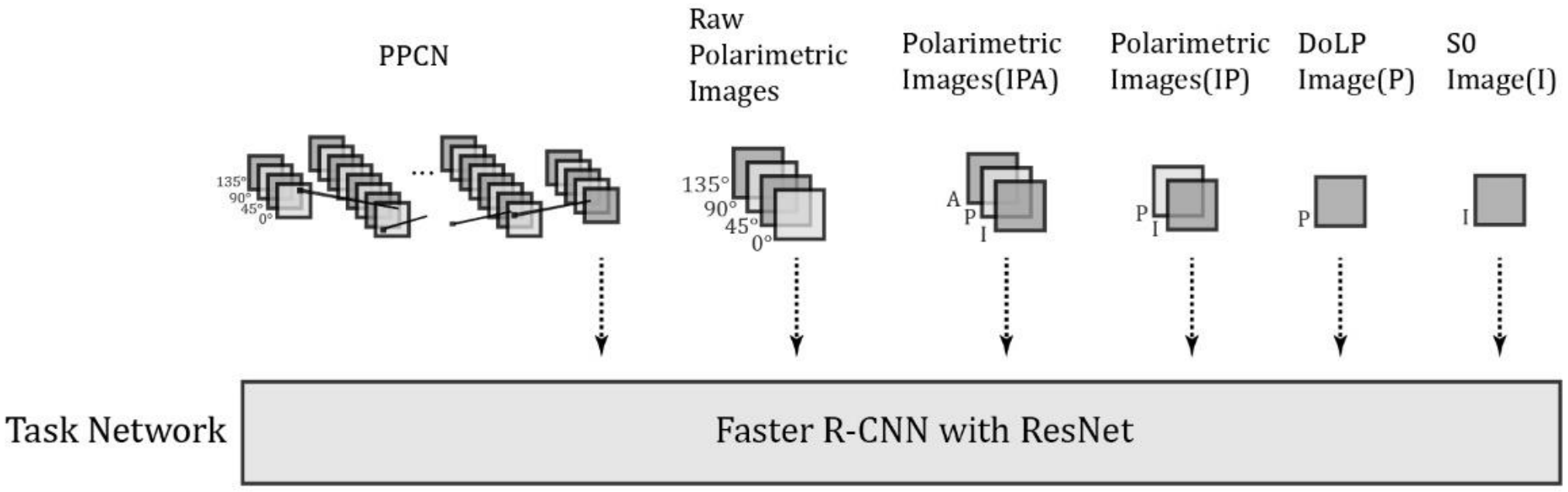

Fig. 7. Different input strategies in experiment.

where I is $S_0$image, A is AoP image, and P is DoLP image. All the polarization-parametric images are normalized to [0,1] according to their respective value ranges.

For each of the input strategies and PPCN, we trained 80 epochs on training set, and then tested on the validation set. To evaluate the performance fully, the IOU-vs-mAP curves of each model were plotted (Fig. 8(a)). We also provided the IOU-vs-AP curves for the two catalogs of the target respectively (Fig. 8(b) and 8(c)).

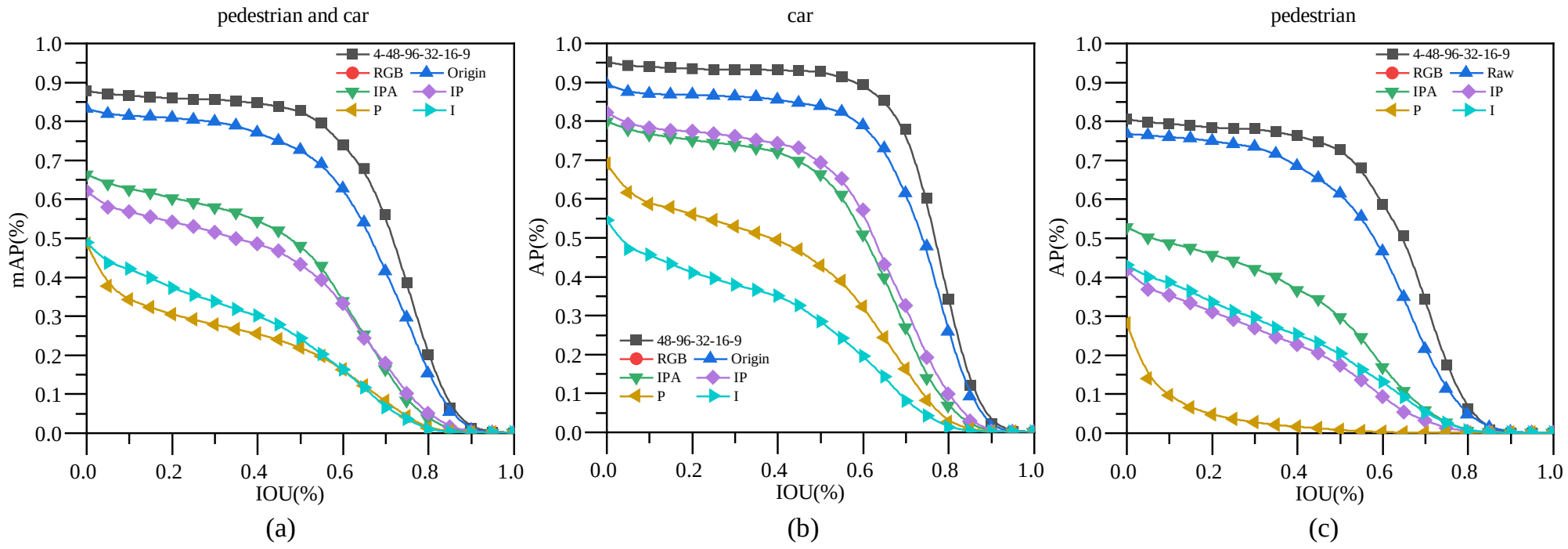

Fig. 8. Curves of IOU-vs-mAP and IOU-vs-AP.(a) IOU-vs-mAP curve for all samples. (b)IOU-vs-AP curve for car. (c) IOU-vs-AP curve for pedestrian.

The results in Fig. 8 prove that the proposed framework is better than all the other input strategies in all cases above. It also shows that different types of targets need to be described by different polarization parameters. For car detection, DoLP(P in Fig. 8(b)) is much better than $S_0$(I in Fig. 8(b)), but for pedestrian detection, DoLP(P in Fig. 8(c)) is the last parameter you should choose. As a counterpart, for the targets that the traditional polarization parameters cannot describe effectively, such as the pedestrian, our framework can construct suitable polarization parameters, which can greatly improve the performance of the task network.

We set the IOU threshold to 0.5, which is commonly used in most cases, and get the mAP and APs shown in Table 5.

**Table 5. Detection Results on Test Set (IOU = 0.5)[a]**

| | Backbone | mAP (%) | Car AP(%) | Pedestrian AP(%) |
|---|---|---|---|---|
| 4-48-96-32-16-9 | ResNet-50 | **0.827** | **0.927** | **0.727** |
| $I_0$, $I_{45}$, $I_{90}$, $I_{135}$ | ResNet-50 | 0.726 | 0.837 | 0.614 |
| $S_0$, P, A | ResNet-50 | 0.480 | 0.663 | 0.296 |
| $S_0$, P | ResNet-50 | 0.433 | 0.693 | 0.172 |
| P | ResNet-50 | 0.217 | 0.427 | 0.007 |
| $S_0$ | ResNet-50 | 0.243 | 0.283 | 0.202 |

[a]GPU=4×1080ti, batch size= 2 per GPU, learning rate = 0.01, input resolution=1280×1024 pixel.

It shows that except our method, raw polarimetric images( $I_0$, $I_{45}$, $I_{90}$, $I_{135}$) can achieve better performance than others. By contrast, our method leads to 10.8% improvement in car detection and 18.4% improvement in pedestrian detection.

Compared with other networks, the size of PPCN-based framework is bigger. However, this does not mean that we can achieve the similar effect just by increasing the size of the task network. To prove that, we changed the backbone of the task network from 50-layer ResNet to 101-layer ResNet. Then we tested with raw polarimetric images, which performed much closer to PPCN than others in the previous experiment. The result is shown in Table 6.

**Table 6. Detection Results with raw polarimetric images using ResNet101 (IOU = 0.5)**

| | Backbone | mAP (%) | Car AP(%) | Pedestrian AP(%) |
|---|---|---|---|---|
| $I_0$,$I_{45}$,$I_{90}$,$I_{135}$ | ResNet101 | 0.712 | 0.825 | 0.599 |

[a] GPU=4×1080ti, batch size= 2 per GPU, learning rate = 0.01, input resolution=1280×1024 pixel.

The result shows that the bigger task network could not provide help in this case, which means that PPCN improves performance in a different way and cannot be replaced with traditional CNN structures.

Further, by putting RGB images into the task network, we studied the difference between polarimetric images and RGB images (Table 7).

**Table 7. Detection Results on Test Set using RGB and PPCN (IOU = 0.5)[a]**

| | Backbone | mAP (%) | Car AP(%) | Pedestrian AP(%) |
|---|---|---|---|---|
| 4-48-96-32-16-9 | ResNet-50 | **0.827** | **0.927** | 0.727 |
| RGB | ResNet-50 | 0.821 | 0.881 | **0.760** |

[a]GPU=4×1080ti, batch size= 2 per GPU, learning rate = 0.01, input resolution=1280×1024 pixel.

Obviously, RGB images perform better on pedestrian detection and polarimetric images perform better on car detection. We also noticed that, except our method, all the other input strategies of polarization images could not achieve approximative performance compared with RGB images. See the appendix for selected examples of detection results.

In addition, we have released the source code for this work publicly [27].

## 5. Conclusion

CNN is a powerful tool for image processing. But the previous networks are not designed for polarimetric images. As lacking of pixel-wise operations, those networks can barely take full advantage of polarization information. Limited by structure, it is inefficient to reconstruct existing CNNs to fix that problem. PPCN is an elegant way to provide the ability to CNNs. Adding PPCN to existing networks can improve the performance without any modification. Moreover, PPCN and task network form an end-to-end framework which is adaptable to most of polarimetric vison tasks. It accepts the raw polarimetric images, automatically constructs the best polarimetric expressions for the data, and then completes a certain vision task. Besides, the output images of PPCN are produced by a certain polarization characterization model learned from dataset. The output images can help us better understand target characteristics and how the computers "see" the polarimetric characteristics.

## Appendix

### *A.1. Car detection results*

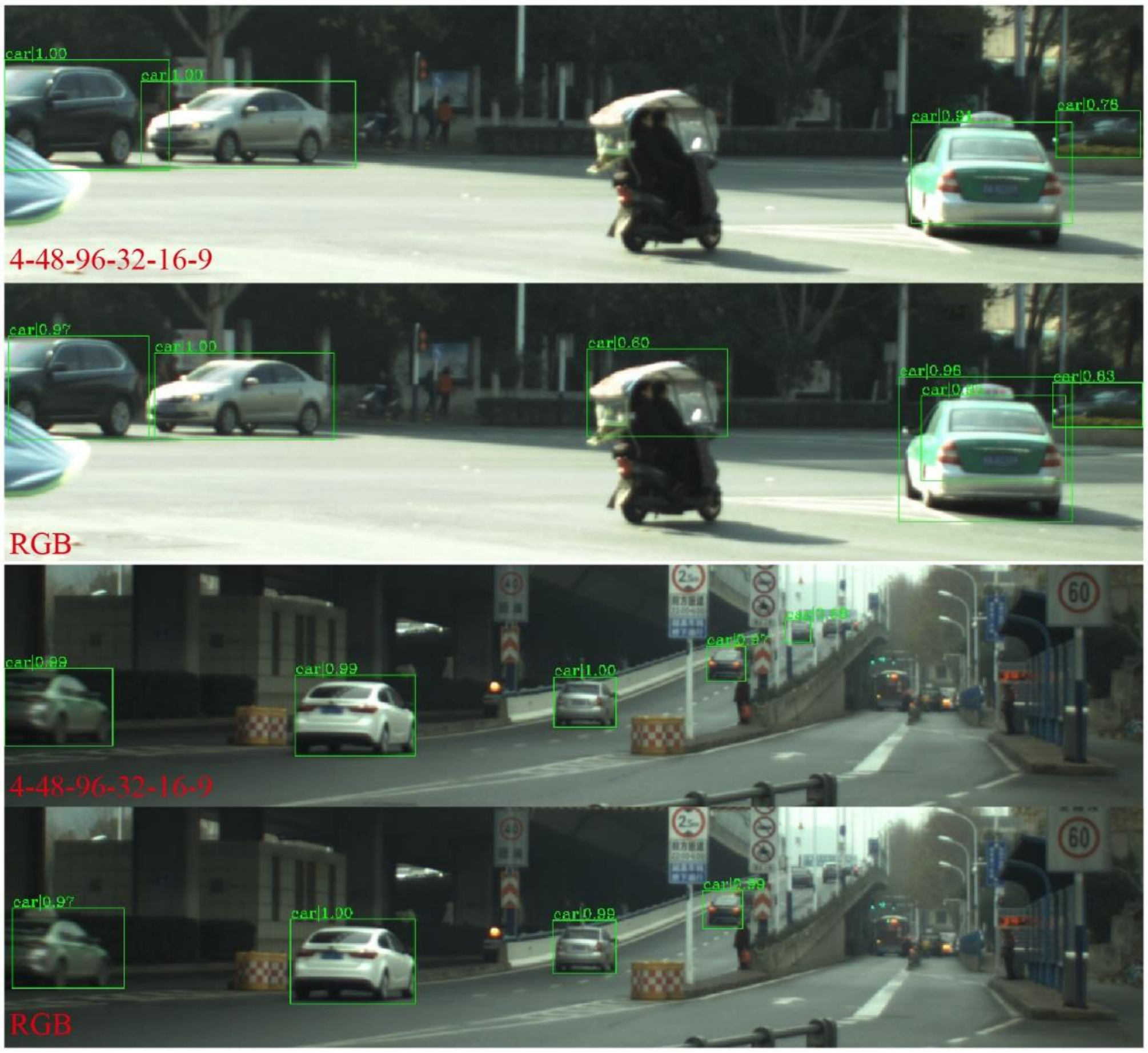

Fig. 9. Selected examples of car detection results of the methods shown in Table 7.

### A.2. Pedestrian detection results

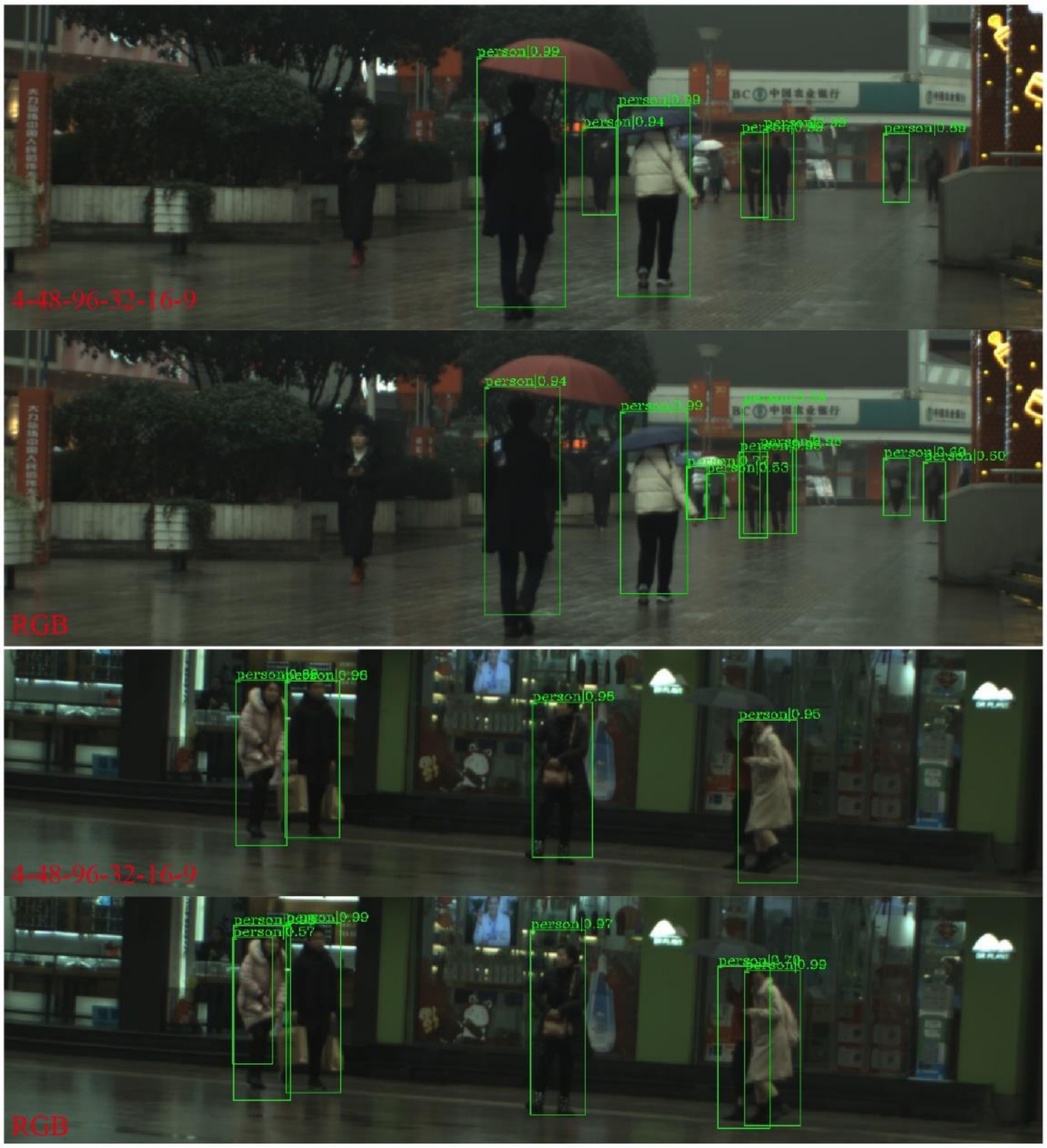

Fig. 10. Selected examples of pedestrian detection results of the methods shown in Table 7.